Shedding light on the Polymer's Identity: Microplastic Detection and Identification through Nile Red Staining and Multispectral Imaging (FIMAP)

By Derek Ho (dkho@upenn.edu)[1*] & Haotian Feng (haotian.feng@ucsf.edu)[2]

1 – Department of Mechanical Engineering and Applied Mechanics, University of Pennsylvania, Philadelphia, Pennsylvania 19103, United States;
Department of Biological Systems Engineering, University of Wisconsin-Madison, Madison, Wisconsin 53706, United States;
2 – Dept. of Radiation Oncology, University of California-San Francisco, San Francisco;
Department of Mechanical Engineering, University of Wisconsin-Madison
\* Corresponding author

**Abstract:**

The widespread distribution of microplastics (MPs) in the environment presents significant challenges for their detection and identification. Fluorescence imaging has emerged as a promising technique for enhancing the detectability of plastic particles and enabling accurate classification based on fluorescence behavior. However, conventional image segmentation techniques for fluorescent particles face several limitations, including poor signal-to-noise ratio, inconsistent illumination, particle thresholding difficulties, and false positives from natural organic matter (NOM). To overcome these challenges, this study introduces the Fluorescence Imaging for Microplastic Analysis Platform (FIMAP), a retrofitted multispectral camera equipped with four distinct optical filters and excited at five different wavelengths. We demonstrate that FIMAP enables comprehensive characterization of the fluorescence behavior of ten Nile Red-stained MPs (HDPE, LDPE, PP, PS, EPS, ABS, PVC, PC, PET, PA) while effectively excluding NOM. This is achieved through K-means clustering for robust particle segmentation (Intersection over Union = 0.877) and a 20-dimensional color coordinate multivariate nearest neighbor approach for MP classification (>3.14 mm), yielding a precision of 90%, accuracy of 90%, recall of 100%, and an F1 score of 94.7%. Among the ten MPs, only PS was occasionally misclassified as its expanded form (EPS). For smaller MPs (35–104 µm), classification accuracy declined, likely due to reduced fluorescent stain sorption, fewer detectable pixels, and camera instability. However, integrating FIMAP with higher-magnification instruments, such as a microscope, may enhance MP identification accuracy. In summary, FIMAP introduces an automated, high-throughput framework for the comprehensive detection and classification of MPs across large environmental sample volumes.

1. **<u>Introduction</u>**

With an estimated quantity of 51 trillion microplastics (MPs, defined as <5mm in size) in the sea, 500 times more than the stars in the galaxy [1], MPs have emerged as an important environmental pollutant, prompting extensive research into their detection, quantification, and classification. These MPs have the potential to induce adverse effects, including oxidative stress, reproductive toxicity, metabolic disorders, and neurotoxicity [2], underscoring the necessity for further investigation. Notably, MPs have been discovered in human placenta [3], blood [4], and heart plaques [5], highlighting the urgency to determine their quantities and transport routes to prevent MP contamination. Long-term monitoring data are required to gain reliable information on MP loading and distribution [6], to understand the extent and degree of the MPs contamination before efforts to alleviate this plastic pollution problem, rendering the need for MP methodologies that are fast and have high throughput suitable for environmental sampling. Quantifying MPs is yet to be standardized and involves various methodologies, including sampling and treatment of environmental matrices before pre-concentration through filtration [7]. Subsequently, high-resolution imaging instruments, such as microscopes, have been utilized to manually sort and count MPs based on their size, color, geometric morphology, and textural features [8]. However, these manual sorting procedures are prone to human error, time-intensive, and cumbersome. This limitation underscores the imperative for automated techniques to expedite the sorting and counting process.

To develop high-throughput methods, researchers are utilizing computer vision and machine learning approaches, leveraging algorithms to classify MPs based on their physical properties. Notably, cascading classifiers integrating geometric, color, and textural features have demonstrated efficacy in differentiating MP particles based on their morphology, significantly reducing sorting time compared to manual methods [9]. While these sorting techniques provide insights into the scale of the MPs prevalence, they are inadequate to classify MPs by their polymer species. As such, validation using spectroscopic analysis, such as Raman and Fourier-transform infrared (FTIR) spectroscopy, is essential to confirm the identity of MPs [8]. However, limited access to spectroscopic machines due to cost and expertise constraints necessitates the development of cost-effective methods for MPs identification and quantification.

Another potential approach involves leveraging the fluorescence behavior of MPs for detection and identification. Researchers have employed machine learning to detect and classify MPs (polyamide (PA), polyethylene (PE), polyethylene terephthalate (PET), Poly(methyl methacrylate) (PMA), polypropylene (PP)) based on their auto-fluorescence, albeit limited to single-particle analysis [10]. To enhance detection capabilities beyond the autofluorescence of MPs, the hydrophobic dye Nile Red (NR) is commonly used to stain plastics, inducing fluorescence upon excitation. Fluorescence imaging comprises an excitation light source, the fluorescent sample (NR-stained MPs), and an imaging device (microscope/camera) for fluorescent detection. Upon exposure to light spanning from ultraviolet (UV, 254nm) to green (500 nm), the NR fluorophore fluoresces upon returning to the ground state, exhibiting a longer wavelength in a process termed the *Stokes shift* [11]. In previous work, the complex interaction between the NR fluorophore, its carrier solvent, and ten polymers (PP, high-density polyethylene (HDPE), low-density polyethylene (LDPE), polystyrene (PS), expanded PS (EPS), acrylonitrile butadiene styrene (ABS), polycarbonate (PC), polyvinyl chloride (PVC), PET, and PA) was investigated [12]. This study identified the optimal carrier solvent—25% (v/v) acetone—to reliably elicit the polarity-dependent fluorescence of polymers, enabling improved classification [12].

Despite this chemical standardization to induce strong and appropriate fluorescence, the detection and the classification of these polarity-dependent *Stokes shift*s for MPs identification has remained problematic. Presently, studies rely on using global thresholds in black and white to select regions of interest (ROI) of MPs across the image. This step is usually performed by establishing brightness thresholds with arbitrary units [13] or fluorescence intensity (FI-%) [14], in image processing software, like ImageJ [18, 19, 20, 21, 22]. Using ImageJ, several semi-automated MPs detection and quantification tools, such as MP-(Visual Analysis Tool) VAT [15], MP-VAT 2.0 [18], and Particle Detection Model (PDM) [19] have been developed. As accurate as some of the detection tools can be, such as the PDM, which is able to distinguish natural organic matter (NOM), such as chitin, cotton, flax, hemp, silk, wood, and wool, from MPs with 95.8% accuracy, these pre-selected thresholds are not arbitrary and may inadvertently exclude low fluorescing MPs (highly crystalline [14] or small MPs [13]), thereby reducing the detectability of MPs. Whereas, setting high thresholds to include all fluorescent particles, including NOM, results in MP overestimation and mislabeling [20]. Another method of thresholding that has been experimented with is adaptive thresholding using deep learning with a tool called MP-Net to distinguish pixels that represent MPs from the background pixels by being trained on the selection criteria previously used by researchers. Even though MP-NET was found to have high precision and sensitivity, as evident from the highest F1-score (0.736) and Intersection over Union (IoU = 0.617) [20], the image segmentation processing is very computationally intensive compared to MP-VAT, its derivatives, and MP-VAT 2.0. Furthermore, MP-NET is trained with human selection criteria, making the training data subjective, resulting in biased data selection. Therefore, improved automated image segmentation tools are indispensable for objectivity, inclusivity, and reproducibility of MP detection in the environment.

In addition, classifying the detected fluorescence behavior for identification has remained challenging. Unlike multispectral cameras that detect changes in wavelengths (nm), most commercial imaging devices (cameras, microscopes, smartphones) are fitted with Bayer filters to capture color in different combinations of red, green, and blue (0-255) within the RGB color space. Though faithful in interpreting transmitted light in 16.7 million colors [21], the Bayer filter is not well-suited for representing the relationship between the polarity-induced fluorescence behavior of the MPs. As such, researchers have attempted to circumvent this issue by quantifying these *Stokes shift* with different formulaic representations of RGB channels, such as (Red+Green/Red) or (Red-Blue/Red+Blue), or layered combinations of these "polarity index" to represent the polarity-induced *Stokes shift* of the MPs for differentiation [2, 3, 6]. Recently, Meyers et al. (2022) proposed a classification method, Particle Identification Method (PIM), that used machine learning to develop a decision tree model to classify the statistics of the RGB colors of the MPs illuminated across three different excitation wavelengths [19]. Their method achieved 88.1% accuracy in identifying various plastic types such as PE, PP, PET, PA, polyurethane (PUR), PS, and PVC [14]. However, some degree of misclassification remains due to metamerism, where two plastics appear similar under one lighting condition but different under another lighting condition, notably between PP and PE, PUR and PA, and PE and PA. Therefore, there exists a strong need for the development of computational tools that facilitate accurate recognition and segmentation of MP in images that could simultaneously classify the select MPs for quantification and use fluorescence colors to classify MPs.

To address this need, we propose an automated framework - Fluorescence Imaging for Microplastic Analysis Platform (FIMAP), which integrates image data from a multispectral fluorescent imaging device with machine learning to optimize (i) image segmentation and (ii) identification of fluorescent particles (PP, HDPE, LDPE, PS, EPS, ABS, PC, PVC, PET, PA, and natural organic matter (NOM)), mitigating misclassification due to metamerism. Furthermore, we investigated how particle size affect the fluorescence detection and identification of NR-stained MPs. Through these inquiries, this study aims to develop FIMAP as a reliable method for simultaneous detection and identification of MPs that is both rapid, simple, and cost-effective, and capable of large-scale environmental sampling.

## 2. Material and Methods

### 2.1 Reference materials and staining procedure for micro-sized MPs

We investigated the ten most common non-colored polymer types: HDPE, LDPE, PP, PS, EPS, ABS, PC, PVC, PET, and PA. These polymers were selected and arranged in order of increasing polarity, as determined by their dielectric constants [24] (see S1 for details). Natural colored HDPE, LDPE, PP, and ABS virgin pellets were sourced from LNS Technologies, while PS came from Sigma Aldrich (#331651), and PA was obtained from GUM Waxed Floss. EPS, PC, PVC, and PET were sourced from post-consumer products such as polystyrene packing peanuts, polycarbonate acrylic glass sheets, PVC pipes, and Dawn Ultra dish soap PET bottles. The NOM used included cotton (from Coats & Clark), chitin (extracted from store-bought eggshells and cut fingernails) and wood (shaved from Jack pine soft wood). Due to the irregular shapes of the microplastic particles obtained from post-consumer goods, their dimensions ranged widely, from as thin as 0.96 mm to 2.09 cm (as seen in PA thread). On average, the dimensions of the studied particles ranged from 3.14 mm (ABS) to 5.42 mm (PET), except PA (10.93 mm) due to its high length-to-breadth ratio of its thread.

To produce micro-sized MPs, we utilized the SPEX SamplePrep 6875D Freezer/mill, employing liquid nitrogen to cryogrind the MPs. The plastics were soaked overnight in water and then in ethanol to ensure thorough cleaning. Polymer samples weighing 2 g were precisely measured and loaded into polycarbonate vials. Subsequently, 8 samples were simultaneously chilled and ground, with 2 vials dedicated to each plastic to prevent cross-contamination. Conducting 5 replicates for each duration and plastic type resulted in 10 g of material for each size and type. Overall, the resulting MPs for our study ranged from 3140 to 5400μm for the larger MPs and 35 to 105μm after cryogrinding.

### 2.2 Pretreatment and NR staining of NOM and spiked biosludge samples

To investigate the ability to differentiate MPs and NOM, we used activated sludge (solid faction that remains after secondary wastewater treatment) obtained from a large municipal wastewater treatment plant in Madison, WI (influent flow: ~45 MGD, two-stage anaerobic digestions for solids processing producing Class A Biosolids), as a surrogate for NOM [25]. The sludge sample (4-5% total solids) were collected using 1 L glass jars and stored at 4°C until further processing. It should be noted that wet sludge is used for analysis as dried sludge tends to contain hardened and clumped clay-like material, which is not easily digested [97].

To prepare the environmental samples for laboratory analysis, thorough stirring was performed to ensure homogeneity before portioning 0.5 g of the sludge into separate 50 mL beakers. One beaker served as a blank control, while the others were spiked separately with 10 different MPs (PP, HDPE, LDPE, EPS, PS, ABS, PC, PVC, PET, PA). The blank samples demonstrated that there was no contamination from MPs originating from the laboratory equipment and procedures. These prepared samples were subjected to Fenton's reagent pretreatment [43]. The pretreatment method for Fenton oxidation was adopted from Masura et al. (2015) [7]. In this adaptation, we specifically excluded density separation through a salt solution to ensure the inclusion of MPs of all densities. Ferrous sulfate heptahydrate ($FeSO_4 \cdot 7H_2O$) (7782-63-0), concentrated sulfuric acid ($H_2SO_4$) (7664-93-9), Whatman 47 mm 2.5μm Grade 42 filter paper were obtained from Fischer Scientific company LLC. Hydrogen peroxide ($H_2O_2$) 30% (v/v) (7732-18-5) was obtained from Santa Cruz Biotechnology. The Fe(II) stock solution was prepared by adding 7.5 g of $FeSO_4 \cdot 7H_2O$ to 500 mL of water and 3 mL of concentrated sulfuric acid [7]. After applying Fenton oxidation to the sample (MPs + NOM) in a water bath set at 70°C for 30 minutes, with periodic agitation every 10 minutes to ensure mixing, the samples were filtered out of the Fenton solution onto the folded 90 mm 2.5μm pore size (Thermofisher Scientific, 09–804–24C) filter paper with a membrane vacuum filtration apparatus (frit size 47 mm), as illustrated in Fig S1. DI water was used to wash off any residual Fenton reagent. This method, utilizing a single oversized filter to contain the MPs sample after Fenton treatment for NR staining, ensures minimal loss of samples during filtration processing. Subsequently, the oversized 90 mm filter paper was then fitted into the bottom of a 60 mm petri culture dish (Inner D: 49.75 mm) with the aid of a 46 mm crucible lid as a stamp to ensure that the filtered area would be completely and evenly. In each of these dishes, 15 mL of NR solution [10 μg/mL] in acetone/water (25% v/v) was added and placed into a water bath (70°C) for 30 minutes, with periodic agitation every 10 minutes to ensure even staining [12]. After staining, the filter paper was lifted and placed back onto the vacuum filter to drain any residual NR solution, resulting in MPs treated with Fenton-oxidation and NR stained within a single filter paper. To evaluate the accuracy of the segmentation of non-overlapping particles, we arranged the particles with tweezers on a 47 mm filter paper for fluorescence imaging.

**2.3 Illumination and Imaging system of FIMAP**

The FIMAP consists of an illumination unit and an imaging system housed in a prefabricated dark room (See Fig S1). The illumination device was designed to have even increments (~50 nm) of excitation wavelengths, specifically 265, 310, 365, 405 and 450nm (See Fig S1 for details). This range of wavelengths was based on preliminary observations that longer wavelengths (e.g., 525 nm) reduced the ability to image NR fluorescence and increased the presence of false positives from NOM [26]. To achieve accurate color capture of the sample, a 45°/0° measurement geometry was employed, where the sample was illuminated at a 45° angle, facilitating diffused reflection for precise color representation.

The intensity and luminosity of the fluorescent illumination for each wavelength were adjusted using a MOSFET switch connected to an Arduino. Since each wavelength had a different power output, the imaging system was calibrated by adjusting the setting of the Arduino to achieve similar fluorescence intensities across different lighting/filter conditions. The LEDs were measured using a luminosity meter on the same plane as the filter paper containing samples.

To image the fluorescent particles on the filter paper, we positioned a 26.2-megapixel full-frame Canon RP Digital Single Lens Reflex (DSLR) camera directly above the center of the MPs, as illustrated in Fig S1. This camera was paired with a Canon RF 100 mm F2.8 macro lens, chosen for its short effective minimum focus distance of 9.2 cm and its ability to accurately reproduce images at a 1.4x magnification ratio. The camera was set to F2.8 at ISO 100 to achieve a shallow depth of field, and minimal noise. The use of this full frame camera setup allowed an increased field of view (FOV), capturing the 47 mm filter paper in its entirety at a resolution of 11.65 μm/pixel. Accounting for potential interference due to noise, this allows for the conservative detection of MPs sized 36.4μm (3 pixels). Additionally, the vignetting effect commonly observed around the edges of images was minimized by capturing high-quality images in a 1:1 aspect ratio. To achieve more accurate color representation in the camera, especially under various lighting conditions, the color temperature was set to 2500 K. This setting preserved the warm tones of the fluorescent MPs amidst the cool tones of the excitation wavelengths [27].

To ensure compatibility across different camera makes and models, we calculated the absolute exposure value (EV) and the luminous exposure value [26] to standardize the amount of light entering the imaging device, as outlined in Table S1. We also enhanced the dynamic range of the captured images through bracket photography or High Dynamic Range (HDR) techniques. This step involved taking multiple photos at different exposure settings (under/over EV±1) and using the camera's built-in processing function to combine them. This resulted in extended highlights and shadows, improving the overall image quality.

To create optimal imaging conditions and enhanced plastic differentiation accuracy, we mounted a series of Cokin optical filters (Ref 001 Yellow, 002 Orange, 003 Red, 004 Green) to filter out extraneous light from the LEDs. These filters help selectively transmit filtered wavelengths of light allowing for more accurate differentiation among plastics, even when they exhibit similar fluorescence colors due to metamerism. Metamerism refers to the situation where materials show the same spectra under specific lighting conditions but differ under altered lighting. Together the five excitation wavelengths (265, 310,

365, 405 and 450 nm with a beam spread of 100-120°) and four optical filters with known spectrum transmission curves, allow the selective capturing of discrete fluorescence behavior filtered across multiple excitation wavelengths, resulting in 20 unique multispectral transmitted color data, as illustrated in Fig 1.

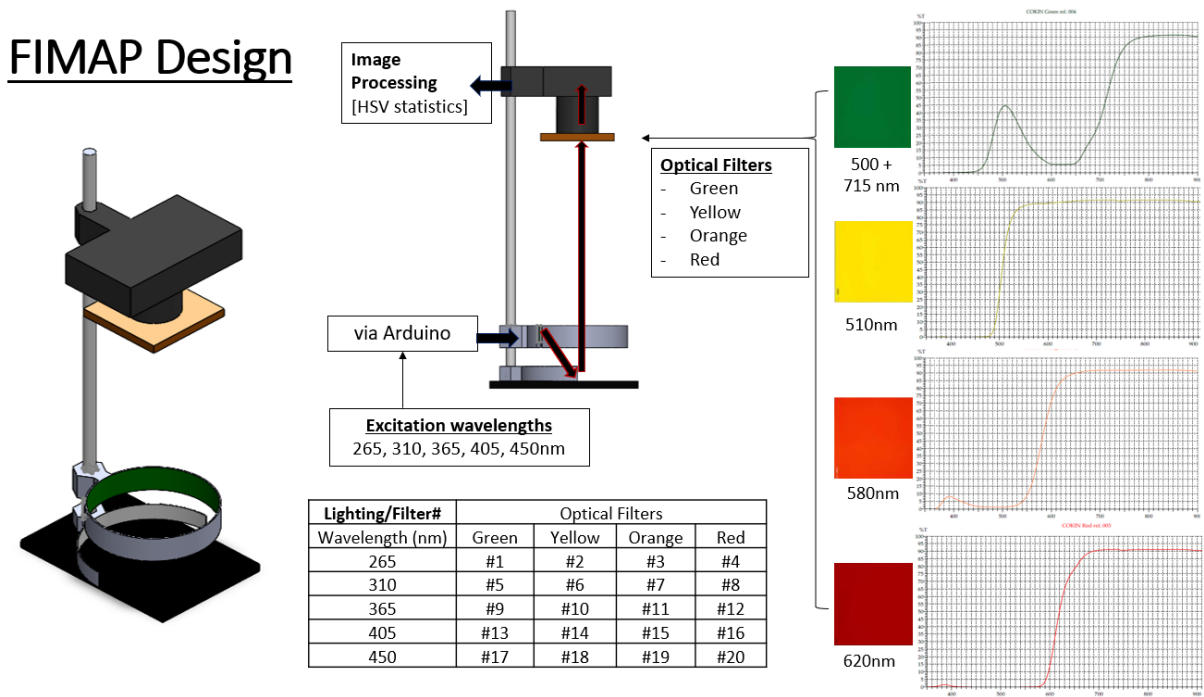

Figure 1: Schematic of FIMAP (265, 310, 365, 405 and 450nm) with the spectral transmission curves of the Cokin filters (green/yellow/orange/red).

**2.4 Image Segmentation of MPs**

We applied k-means clustering to segment the fluorescent particles in the captured images with Matlab. We found that the YCbCr color space performed better in selecting for uneven fluorescence across particles compared to the RGB color space. For example, some plastics like translucent PET exhibit greater fluorescence at their edges than at their centers, posing a challenge for selecting the entire ROI using the RGB color space. In this case, YCbCr, which separates color into luminance (Y) and chrominance (Cb and Cr) components [28], outperformed RGB in its ability to select PET in its entirety. Despite the issue of potentially over-representing particle size, we utilized the YCbCr-based segmentation method to ensure comprehensive plastic detection, albeit with some instances of enlarged ROIs.

To ensure the detection of all particles, we evaluated the rates and accuracy of segmentation across all 20 lighting/filter conditions. Among the 20 different conditions, only the 310 nm/orange filter (#12) and 450 nm/red filter (#20) exhibited the detection of all 10 MPs (See Table S2). Of the two, lighting/filter #12 was chosen because it elicited a stronger fluorescence response from all 10 MPs, especially PA. This lighting/filter condition was used to generate the binary mask "Mask-12" template to be applied across all lighting/filter conditions to extract features such as location, area, morphology, and HSV color properties. The Mask-12 layer was obtained by doubling the exposure value settings of #12 with 'imfuse' function in Matlab to fill any "holes" formed caused surface reflection with the lights. Mask-12, containing the location and size of fluorescent particles, could be uniformly applied across the aligned image stack, facilitating the extraction of statistical and HSV (Hue, Saturation, and Value) color features of the particles of interest across 20 dimensions, resulting in a 20-by-2 vector containing mean and standard deviation. We choose HSV color space as it proved ideal for portraying fluorescent colors [12]. In HSV, hues are arranged on a 360° color wheel, with primary colors at 0°, 120°, and 240°, while saturation and value determine richness and brightness, respectively. Due to its intuitive arrangement of colors, this color space provides a more appropriate correlation with the polarity-induced *Stokes shift* of the plastics.

**2.5 MPs classification with Multispectral data**

After employing this material-specific spectral fingerprint, we further conducted a nearest neighbor search (NNS) to match the 20 color coordinates with those of other materials. This method entails comparing the nearest points in a dataset to a given query point and assigning points based on their proximity to cluster centers [29]. Accounting for the color reflection variance introduced by particle sizes change, we utilize the Mahalanobis distance to measure the distance between the corresponding two points [29].

## 3. Results

### 3.1 MPs segmentation using k-means

Current particle segmentation tools using ImageJ are limited to global thresholding (Grayscale/color). As Nel et al (2021) has mentioned, selecting this threshold is not arbitrary. When using grayscale values, like FI (%) the color data was suppressed onto a binary scale, and limits the selection of low-fluorescent MPs that may be misconstrued with NOM [13]. Color thresholding of non-fluorescent MPs has been performed as well. Vermeiren et al (2020) processed MPs through color thresholding between these parameters - Hue [0–43], Saturation [0–255], and brightness [160–255] on a scale of 0–255 to select for orange and red PE, PP, PS, EPS, PET, PA and PVC [30].

Instead of relying on global thresholding with fixed value or dynamic value like Otsu, our study demonstrated a significantly better performance with k-means clustering (k=3). The k-means clustering segments fluorescent MPs from the background, resulting in designated ROIs. Three k-clusters were used to distinguish the (1) fluorescent particles from (2) the background and (3) shadows cast by the ROI. K-means clustering operates in a multi-dimensional feature space, allowing the analysis of features such as color, intensity, and spatial information concurrently, leading to more nuanced objective segmentation results (See Fig 2). For instance, in Fig S2, where the same MPs were processed using ImageJ, a manual threshold between 9% [54,255] and 10% [48,255] had to be set to achieve a similar level of image segmentation as the proposed K-means clustering method. This improvement is attributed to K-means clustering's ability to optimize cluster centroids based on an objective criterion, such as minimizing the within-cluster sum of squares [16, 17]. This data-driven process begins by assigning pixels to clusters and updates the centroids of the clusters by iteratively computing the mean of all pixels assigned to each cluster until centroids reach a state of equilibrium and stabilize. Upon convergence, pixels undergo segmentation into distinct groups [31]. As a result, particles are segmented more consistently across varying fluorescence intensities and imaging conditions. Unlike manual thresholding, which relies on fixed intensity cutoffs that may not generalize well across different samples, K-means clustering dynamically adjusts to pixel value distributions, reducing the risk of misclassification and enhancing the robustness of MP detection.

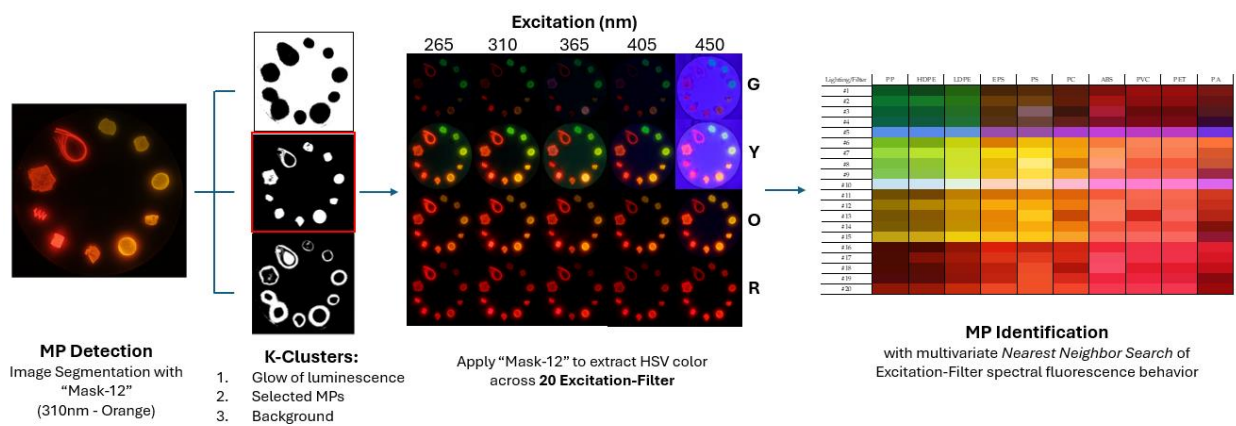

**Figure 2:** Flowchart illustrating the detection and identification of MPs using FIMAP. The process begins with detection via K-means clustering (k = 3), followed by the selection of cluster #2 for segmentation across 20 excitation-filter conditions. HSV color features are then extracted, and a 20-coordinate multivariate Nearest Neighbor search is applied to identify the MPs.

One major limitation to adopting fluorescence imaging for environmental sampling was the occurrence of false positives caused by the fluorescence of NR-stained NOM [32]. To assess the effectiveness of this detection model, we spiked a mixture of 10 MPs and 3 NOMs (chitins and cotton) with 0.5 g of biosolids and treated it with Fenton oxidation, as illustrated in Fig 3. The particles were arranged on another filter paper to minimize interference from MPs in the biosolids, with the MPs placed in the peripheral circle and the NOMs (cotton and chitin) in the center to prevent overlap. Since Fenton oxidation might not have completely removed NOM, we evaluated whether this process could eliminate the detectable fluorescence produced by these NR-stained NOMs. To assess the effectiveness of Fenton oxidation in reducing these false positives, we selected NOM from cotton, chitin (nail) and chitin (egg). From Fig. S3, we observed that Fenton-treated NOM (cotton, chitin and wood) exhibited no detectable fluorescence compared using K-means clustering. This lack of detection of the NOM indicated that Fenton oxidation successfully eliminated the fluorescence of these NOMs, making them easily distinguishable from the MPs in question.

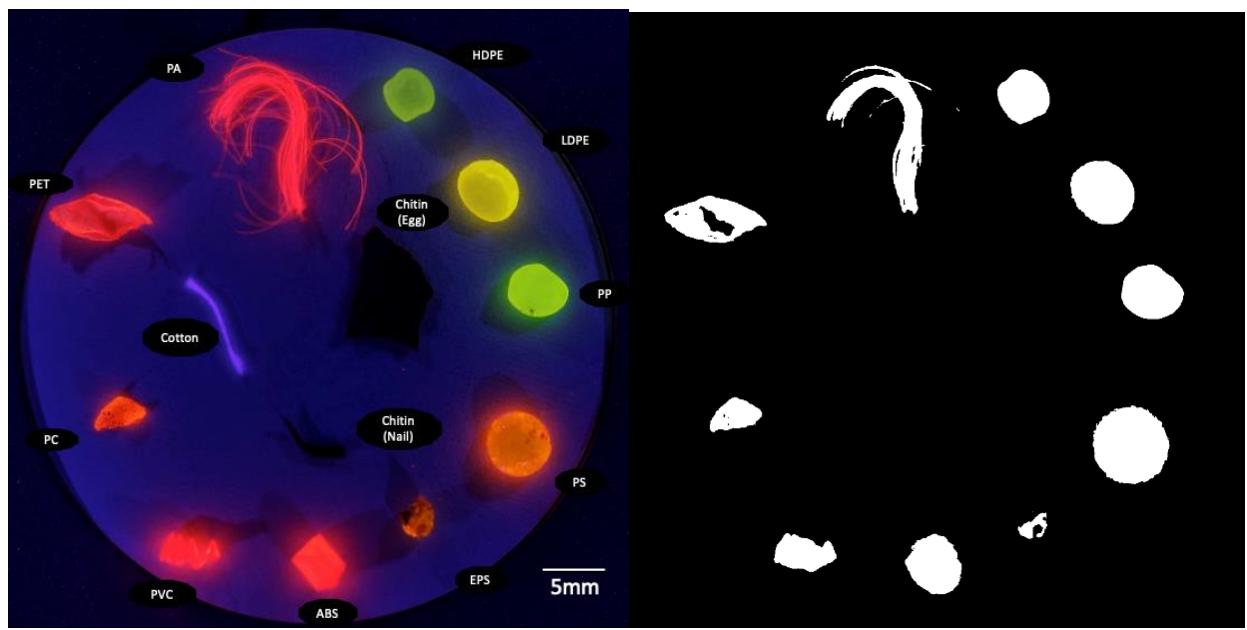

Figure 3: 10 MPs + 3 NOM (Chitin (egg and nail) and cotton) after exposure to biosolid and Fenton oxidation, captured at 265nm with no filter (left); K-means clustering of MPs and NOM, revealing selective detection of only MPs.

**3.2 MPs classification using k-means**

To classify the MPs, the mean and standard deviation of the HSV color data across 20 lighting/filter conditions of the 10 MPs were obtained and used to train the MPs library. The mean HSV color of the 20 conditions are illustrated in Fig 2. The obtained 20 points of "spectral fingerprint", akin to a color constellation, could be used to distinguish the MPs by evaluating the similarities between the color data structures.

In previous work [12], MP classification was assessed through the calculation of color differences ($\Delta E = \sqrt{\Delta H^2 + \Delta S^2 + \Delta V^2}$) using a single lighting condition at 310nm without a filter. While the use of this equation readily differentiated seven MPs (Delta E > 10), it failed to distinguish certain plastics such as ABS, PVC, and PA due to metamerism, causing significant color overlap. To address this limitation, comparing differences across multiple lighting and filter conditions was proposed. This approach leverages the Mahalanobis distance ($D_M$) (Equation 1) [33], which allows for multi-variate statistical assessment of the dissimilarities between datasets while accounting for color variations under varying lighting and filters. The $D_M$ builds upon the concept of z-scores by using a multivariate mean and a multidimensional representation of the covariance matrix instead of traditional mean and standard deviation. This enhancement within multidimensional space. allows for a comprehensive assessment of color variations across multiple dimensions. In our study, we store the Mahalanobis distances of the selected MPs in a distance matrix (in the measure of standard deviations), where each entry indicates the $D_M$ between two materials (See Table 1). A $D_M$ value indicates a high similarity between datasets when it approaches 0 and a low similarity with larger values.

Mahalanobis Distance, $D_M = \sqrt{(x-m)^T C^{-1} (x-m)}$ [Unit: *Standard Deviation*]   (Equation 1)
Where, x = Vector of data
M = Vector of mean values of independent variables
$C^{-1}$ = Inverse Covariance matrix of independent variables
T = Indicates vector should be transposed

Table 1: Distance matrix table of Mahalanobis distance of the 10 virgin MPs (PP, HDPE, LDPE, EPS, PS, PC, ABS, PVC, PET, PA), where values approaching 0 indicates a higher degree of similarity between datasets, while larger values, signify increased levels of dissimilarity. ($D_M$ < 1 are highlighted)

|      | PP | HDPE | LDPE | EPS | PS | ABS | PC | PVC | PET | PA |
|---|---|---|---|---|---|---|---|---|---|---|
| PP | 0 | 1.02 | 2.05 | 2.86 | 12.79 | 23.57 | 14.62 | 22.77 | 8.40 | 8.43 |
| HDPE |  | 0 | 1.93 | 2.59 | 10.58 | 21.15 | 11.15 | 20.19 | 7.54 | 8.03 |
| LDPE |  |  | 0 | 1.63 | 5.97 | 13.73 | 6.60 | 13.52 | 4.38 | 5.45 |
| EPS |  |  |  | 0 | 0.45 | 5.87 | 1.21 | 6.39 | 2.78 | 4.57 |
| PS |  |  |  |  | 0 | 15.18 | 4.80 | 15.84 | 10.97 | 12.77 |
| ABS |  |  |  |  |  | 0 | 13.28 | 1.42 | 7.99 | 14.46 |
| PC |  |  |  |  |  |  | 0 | 14.07 | 10.22 | 11.94 |
| PVC |  |  |  |  |  |  |  | 0 | 8.92 | 13.49 |
| PET |  |  |  |  |  |  |  |  | 0 | 4.57 |
| PA |  |  |  |  |  |  |  |  |  | 0 |

Among the 10 virgin MPs studied, two polymers – PS and its expanded form, EPS, showed the only $D_M$ value (0.45) below 1 standard deviation, indicating a high likelihood of misclassification. This result is understandable considering these plastics share similar chemical characteristics but differ in physical densities. Additionally, other plastic pairs had a $D_M$ values larger than 1 to a maximum of 23.57 standard deviation, indicating significant differentiation and classification ability between them. This is a significant improvement from the previous use of ΔE which is prone to metamerism, allowing the previously indistinguishable plastics to be easily distinguishable (with a $D_M$ value greater than 1 standard deviation), as seen between ABS and PVC ($D_M$ = 1.42), PVC and PA ($D_M$ = 13.49) and ABS and PA ($D_M$ = 14.46) [12].

### 3.3 Evaluation of detection and classification model

To assess the accuracy of our ROI segmentation, we measured the pixel areas of all detected particles. Because the sizes of detected ROIs varied under different lighting and filter conditions (See Table S2), determining the true scale of the MPs compared to the ROI was challenging. Therefore, to reduce bias towards ROI enlargement by using the mean values, we compared the ROI from Mask-12 to the median area across 20 filters. It was observed that MPs like PP, HDPE, LDPE, and PC appeared relatively true to scale (99.9–104.1%). However, some plastics show enlarged ROI percentages, such as PS

(113.8%), compared to their median pixel areas across the 20 conditions. Conversely, certain plastics exhibit smaller detected ROIs, due to sludge accumulation obstructing fluorescence in crevices (e.g., EPS at 46.7%), or variations in plastic thickness (e.g., PET at 92.0%), or undetected fine threads (e.g., PA at 78.6%). Although using the median offers a fair comparison for most plastics, some plastics like ABS (98.5%) and PVC (85%) show smaller ROI values despite apparent enlargement in Fig 2. To offer a more accurate representation of these MPs, comparing with the first quartile values reveals ABS at 117.8% and PVC at 107.9%.

FIMAP demonstrated that it was able to reliably detect the 10 virgin MPs apart from the NOM; resulting in 10 out of 10 true positives (TP), zero false negatives (FN), one false positive (FP) and zero true negative (TN). To evaluate the detection and classification effectiveness of FIMAP, we utilized the evaluation metrics of IoU, accuracy, precision, recall, and F1 score (See Equations 2– 6) [36, 37]. IoU (Equation 2) assesses the degree of overlap between the predicted segmented region and the ground truth region [36, 37]. Accuracy (Equation 3) reflects the overall correctness of the model, precision (Equation 4) measures the accuracy of positive predictions, and recall (Equation 5) gauges the ability of the classification model to correctly identify actual positives [36, 37]. The F1 score (Equation 6) serves as a harmonic mean between precision and recall; with a high value indicating effective identification of correct detection while minimizing false positives [36, 37].

$$IoU = \frac{TP}{(TP+FN+FP)} \quad \text{(Equation 2)}$$
$$Accuracy = \frac{TP}{(TP+FP)} \quad \text{(Equation 3)}$$
$$Precision = \frac{TP+TN}{(TP+TN+FN)} \quad \text{(Equation 4)}$$
$$Recall = \frac{TP}{TP+FN} \quad \text{(Equation 5)}$$
$$F1\ score = \frac{2*Precision*Recall}{(Precision+Recall)} \quad \text{(Equation 6)}$$

Where, TP = True positive – detected and correctly identified MPs [34]
FP = False positive – detected but incorrectly identified MPs
TN = True negative – detected but identified as not MPs
FN = False negative – MPs not detected

To determine the IoU, we compared the pixel area segmented by Mask-12 with the median pixel areas across the 20 conditions (except for ABS and PVC, where the first quartile values were used). The IoU values for the 10 MPs ranged from 0.467 (EPS) to 0.999 (PP), with an average IoU value of 0.877 for all ten MPs (See Table S2), outperforming the best deep-learning-based segmentation techniques of MP-Net (IoU = 0.617) [20]. We observed that FIMAP correctly disregarded NOMs and identified nine of the ten MPs, with only EPS being misclassified as PS (Table 1). In summary, FIMAP demonstrated a precision of 90%, accuracy of 90%, with a recall of 100%, and an F1 score of 94.7% for non-overlapping Fenton-treated NR-stained MPs (>3.14 mm). Compared to studies focusing solely on MP segmentation, FIMAP surpassed deep learning models such as U-Net, which achieved the highest MPs detection accuracy of 69.2% and recall of 88.3%, and MP-Net which had the highest F-1 score of 73.6% [20], as summarized in Table 2. This improved performance can be attributed to the use of K-means clustering instead of conventional global thresholding, highlighting its efficacy in objectively segmenting NR-stained MPs.

In terms of MP classification ability, FIMAP demonstrated an improved ability to distinguish MPs compared to work done by Meyer et al., who achieved 88.1% accuracy in identifying various plastic types such as PE, PP, PET, PA, PUR, PS, and PVC, but with misidentification between PP-PE, PUR-PA, and PE-PA [19]. Our FIMAP setup, with an increased number of 20 lighting/filter conditions compared to three conditions used by Meyer et al. (2022), facilitated accurate and precise differentiation of ten virgin

MPs. Furthermore, our method overcame some misclassifications (PP-PE and PE-PA) that were due to metamerism using the limited lighting/filter conditions.

Table 2: Evaluation of detection and identification models for MPs analysis

| Evaluation Measure | Detection only<br>MP-Net<br>(Best of 12 different deep learning models*) [20] | Identification only<br>Particle Identification Model (PIM) [19] | Detection & Identification<br>Fluorescence Imaging Microplastic Analysis Platform (FIMAP) |
|---|---|---|---|
| Intersection over union (IoU) | 0.617 | - | 0.467 – 0.999<br>Avg: 0.867 |
| Accuracy | 69.2% | 88.1% | 90% |
| Precision | 82.0% | - | 90% |
| Recall | 88.3% | - | 100% |
| F1 Score | 73.6% | - | 94.7% |
|  | *All models uses conventional subjective thresholding techniques | Misidentification between PP-PE, PUR-PA, and PE–PA | Misidentification between PS-EPS |

### 3.4 Evaluation of Particle size

To evaluate the impact of particle size on classification, we conducted NR staining on six cryoground MPs (PP, HDPE, LDPE, PS, ABS, PET) with sizes ranging from 35 to 104 µm, as shown in Fig. 4. The processed images were analyzed using two methods: FIMAP (K-means clustering) and ImageJ (manual black-and-white thresholding), with results illustrated in Fig. 4b & 4c, respectively, and summarized in Table 2. Differences in FI posed challenges in determining an optimal threshold that could reliably detect all MPs while excluding false positives. To address this issue, we applied a minimum particle size filter of 100 pixels and compared the effects of thresholding ranges from 0.15% [27, 255] to 0.08% [53, 255]. As noted by Nel et al. (2021) [13], some MPs, such as PET, have poor detection performance due to their high crystallinity, requiring a higher threshold (0.15%) to fully identify the region of interest (ROI). This limitation also affected K-means clustering, which segmented the ROI into two parts. PET's translucent nature exacerbated this issue, splitting the ROI detection at lower thresholds (e.g., [53, 255]). Conversely, MPs with strong fluorescence characteristics (PP, HDPE, LDPE, PS, and ABS) exhibited visibly larger ROIs at these thresholds, as detailed in Table 3.

While our classification code demonstrates high potential for larger MPs (3140 to 5400µm), its effectiveness diminishes for smaller MPs (35 to 104µm), as indicated by the $D_M$ values for various MP pairs. As illustrated in Table 3, the low $D_M$ values (<1.0) of PP & LDPE, PP & PS, PP & ABS, PP & PET, PS & LDPE, and PS & PET suggest challenges in distinguishing smaller MPs accurately. The diminished classification performance may be attributed to several factors. Firstly, the reduced fluorescence could stem from limited surface area available for NR sorption, thus reducing the fluorophore available for excitation. This variance in fluorescence behavior is highlighted by evaluating the $D_M$ of larger MPs (>3.14 mm) compared to micro-sized MPs (~100µm), where a significant $D_M$ discrepancy between the large and micro-sized MPs ranging between 3.74 (HDPE) to 21.12 (ABS) was observed. Additionally, the camera setup may experience instability and movement during filter changes, compromising the accuracy of image registration for spectral series. This instability is especially pronounced at smaller sizes, where the smallest detectable particle at the pixel level (11.65μm/pixel) and the narrowest detectable particle measures only 3 pixels wide (35μm).

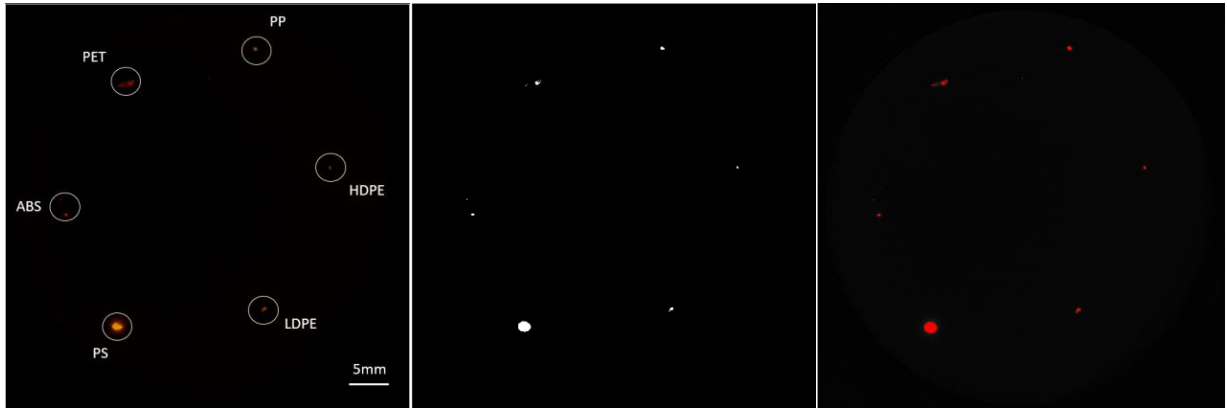

Figure 4a: 6 Micron-sized NR-stained MPs (PP, HDPE, LDPE, PS, ABS, PET – Clockwise from the top) ranging from 35 to 104 μm in size, captured at Mask-12 (310 nm with orange filter),
Figure 4b & c: Comparison between K-means clustering (FIMAP) vs. BnW Thresholding [0.10%] (ImageJ)

Table 2. Particle selection of small MPs (35-104 μm) with FIMAP vs BnW thresholding

| <100 px Polymer | FIMAP K-means clustering (in px) | ImageJ (Manual Thresholding) | | | |
|---|---|---|---|---|---|
| | | Threshold 0.15% [28,255] | Threshold 0.13% [32,255] | Threshold 0.10% [45,255] | Threshold 0.08% [56,255] |
| PP | 1091 | 1803 | 1641 | 1257 | 1052 |
| HDPE | 454 | 748 | 691 | 581 | 463 |
| LDPE | 962 | 1723 | 1528 | 1061 | 834 |
| PS | 11214 | 15227 | 13996 | 11326 | 9696 |
| ABS | 644 /119 | 936/124 | 828/102 | 567/- | 310 /- |
| PET | 1400 (117) | 4418 | 2539 (158) | 1237 (-) | 791 (-) |

Table 3: Distance Summary of Mahalanobis distance of the 6 micro-sized MPs (PP, HDPE, LDPE, PS, ABS, PET) ranging from 35 to 104μm, where values approaching 0 indicates a higher degree of similarity between datasets, while larger values, signify increased levels of dissimilarity. ($D_M$ < 1 are highlighted)

| | PP Small | HDPE Small | LDPE Small | PS Small | ABS Small | PET Small |
|---|---|---|---|---|---|---|
| PP Small | 0 | 1.46 | 0.95 | 0.71 | 0.68 | 0.77 |
| HDPE Small | | 0 | 1.45 | 1.45 | 2.30 | 2.03 |
| LDPE Small | | | 0 | 0.60 | 1.84 | 0.94 |
| PS Small | | | | 0 | 1.29 | 0.40 |
| ABS Small | | | | | 0 | 1.61 |
| PET Small | | | | | | 0 |
| Size (μm) | 47 - 104 | 82 - 94 | 35 - 47 | 82 - 94 | 35 - 47 | 82 - 94 |

3.5 Testing on biosludge samples

To evaluate the applicability of FIMAP in environmental settings, we collected, pretreated and analyzed 1g of biosolid samples containing MPs from WWTPs. Segmentation of the fluorescent particles in biosolids required adjustments, including increasing the number of k-means clusters from three to four to account for biosolid residue on the filter, and the use of lighting/filter condition #20 as the binary mask for segmentation (as seen in Fig 5a). We categorized particles into three size categories: 10, 50, and 100 pixel$^2$ (equivalent to approximately 1365, 6786, and 13752μm$^2$), as illustrated in Table 3. This range was chosen to ensure particles were at least 3x3 pixels to disregard detection of background noise. Using these size ranges, we detected 300 particles larger than 10 pixel$^2$, 160 particles larger than 50 pixel$^2$, and 98 particles larger than 100 pixel$^2$, as illustrated in Fig 5b. The smallest particles detected were approximately 24.6 microns wide (minimum minor axis length). Due to size constraints, we could not validate the identity of these NR-stained particles with FTIR and could only suspect they were plastics. This initial assessments of biosolids suggest that 1 g of sludge may contain around 300 particles/g (>24.6μm) (equivalent to 300,000 particles/kg), which aligns prior estimates of other WWTP exploration, reporting between >1000 particles/kg to 301,400 particles/kg [36].

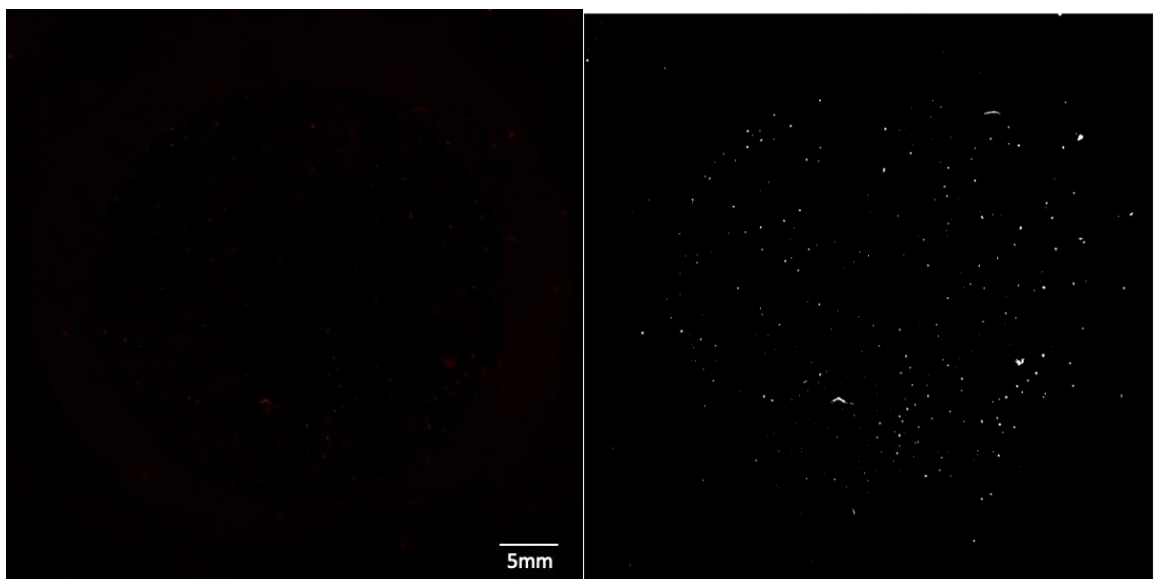

Figure 5a (left): Lighting/filter condition #20 (450nm/Red filter) of Fenton-treated NR-stained of biosolids
Figure 5b (right): Segmentation mask of detailing the detected fluorescent particles in Fig 5a.

4. Discussion

An innovative method, called FIMAP, was developed for the simultaneous detection and classification of MPs by quantifying their fluorescent behavior after pre-treatment and NR staining. FIMAP has demonstrated the ability to accurately and precisely detect and identify the ten most common MPs larger than 3140 μm, with a minor exception in distinguishing EPS from PS, while effectively differentiating MPs from the NOM present in the samples. For smaller MPs (35 to 104 μm), FIMAP could detect the micron-sized particles, though with reduced classification accuracy due to decreased sorption capacity and limitations of the imaging instruments. When tested on biosolids, FIMAP detected 300 particles per gram (>24.6 μm) of suspected MPs, a result consistent with the typical range found in WWTP studies [36]. However, due to reduced identification accuracy at smaller particle sizes and the inability to validate polymer species using FTIR, we could not fully assess FIMAP's polymer identification efficacy.

By using advanced image segmentation techniques, such as K-means clustering, FIMAP improves upon the conventional global thresholding methods observed in current MP analysis systems like MP VAT 1.0 [15], MP-VAT 2.0 [18], and PDM [19]. Unlike global thresholding, this approach objectively isolates fluorescent particles from the background without relying on subjective threshold values, enabling particle selection across multiple dimensions, such as color and intensity. The enhanced detection capabilities of FIMAP can be attributed to its use of the YCbCr color space, which allows for independent analysis of luminance and chrominance components. The adaptation of K-means clustering to this color space improves resistance to variations in lighting, shadows, and background noise, resulting in more nuanced and accurate segmentation. In contrast to the conventional RGB color space used for particle segmentation, YCbCr is less affected by lighting variations, facilitating more effective color segmentation. Notably, the number of K-means clusters may need to be increased from three to four in samples with higher turbidity. This adjustment accounts for additional contaminants, such as biosolids, which can be segmented into a distinct category, improving overall classification accuracy.

When applied to Fenton-treated samples, FIMAP, combined with K-means clustering, objectively detects low-fluorescent MPs while minimizing false positives from NR-quenched NOM, thus enhancing classification accuracy. In addition to improving the objective detection of MPs, FIMAP also facilitates accurate polymer identification using a 20-coordinate spectral fingerprint across different lighting and filter conditions. The material-dependent spectral fingerprints of the extracted HSV color data across all 20 lighting/filter combinations create a unique color spectral fingerprint specific to each MP type. This method replaces traditional RGB channel classification with an HSV color space, where color changes are reflected in the weighted hue-dominant channel, making color distinction easier. The HSV color space, which corresponds directly to the dominant wavelength of light and is independent of saturation and value channels, offers more intuitive and effective fluorescent particle classification. The increased detection across 20 different lighting/filter conditions, combined with nearest neighbor search (NNS) for comparison, overcomes metamerism and distinguishes MPs that were previously challenging to identify. This represents a significant improvement over earlier methods that relied on pseudo-polarity calculations or RGB decision trees to classify MPs based on their fluorescent behavior [14, 19].

Compared to the PIM model, which uses a decision tree based on RGB data statistics, our model is more flexible and can easily accommodate the wide range of polymers found in the environment without the need for re-computation. Expanding the polymer library with FIMAP involves simply adding a 20-coordinate spectral fingerprint for matching during classification, making the method highly adaptable. This flexibility allows for the inclusion of various plastics, taking into account factors such as additives, polymer composition, and weathering, without needing to re-compute a rule-based algorithm.

FIMAP has also proven to be a fast and cost-effective tool for simultaneous MP detection and classification. After the one-hour pre-processing, which includes Fenton oxidation and NR staining, FIMAP can complete the capture of 20 lighting conditions in as little as 15 minutes, process image segmentation and extract features in 30 minutes (for 21 particles), and classify particles by comparing the HSV average and standard deviation library in just 5 minutes. This speed is comparable to Meyer et al.'s FIM, which reportedly takes 55 minutes for 20 samples [19], and faster than current identification methods using µ-FTIR, which can take up to 170 minutes for the analysis of 20 MPs . The main limiting factor is the computational intensity of the segmentation code, which could be expedited by using a dedicated GPU for image processing, potentially speeding up the process by a hundred-fold [37]. This advancement holds the potential to greatly improve large-scale environmental sampling efforts, allowing for the study of spatial and temporal trends in MP transportation, as well as the detection of point sources of pollution that need remediation.

Moreover, the affordability of FIMAP (priced at <$3000) makes it an appealing choice for researchers involved in environmental studies. For instance, with 1g of NR priced at $270 , $146 for 2.5 L of acetone , and $39 for 47 mm filters , thousands of batches (15 ml [10 µg/ml]) can be produced at approximately $0.45 per sample—ten times cheaper than using Meyer et al.'s PIM . This represents significant cost savings compared to spectroscopic tools like FTIR and Raman, which typically range in price from $25,000 to $100,000. This cost-effectiveness of FIMAP can allow for the study of MPs in low-income areas disproportionately affected by plastic pollution, raising awareness of the effects of plastic pollution. FIMAP can also be retrofitted onto other imaging devices, from microscopes to DSLRs and smartphone cameras.

Our current design of FIMAP allows for the detection of MPs as small as 36.4 µm (3 pixels) within a FOV of 47 mm, marking a significant advancement over previous analysis systems such as MP-VAT with 65 µm [15] and PDM's 50 µm [19]. Notably, the FOV of other similar setups, such as the MP-VATs and PDM, is limited to 45 mm and 22 mm at a magnification setting of 10x, respectively [18, 22, 53]. Currently, FIMAP stands out as the first of its kind, allowing for true-to-scale detection and identification of MPs with a large FOV of 47 mm, without the need for image stitching. It is believed that adapting FIMAP to super-resolution microscopes could significantly boost our ability to detect and identify smaller particles, facilitating the comprehensive study of MPs and NPs in a single study [39]. However, to achieve this, more work is needed on color calibration across devices and particle sizes to ensure consistent matching. Before FIMAP can be established as a standalone method, several challenges must be addressed, including the detection of weathered and colored plastics. Additionally, further investigation is needed to evaluate FIMAP's performance with environmental samples and to manage potential particle overlap, which may require reducing sampled volumes in areas with high MP concentrations. The detection of fine threads (e.g., PA) is also limited, raising concerns due to the increasing environmental threat of microfibers. Future efforts could incorporate deep learning techniques, such as convolutional neural networks (CNNs), to enhance accuracy, speed, and particle selection in image processing.

Overall, the development of automated methods can significantly reduce the time required for sampling analysis, decrease subjectivity from human bias, and allow for the standardization of MP analysis in a cost-effective manner. As such, FIMAP holds great promise for enhancing pollution monitoring and assessment, enabling accurate spatial and temporal mapping of MPs in various environments, from flora to fauna. This information can provide policymakers with the data needed to formulate effective, tailored regulations and policies locally, regionally, and internationally. The identification tool can also pinpoint hotspots for MP pollution, offering insights to better inform waste management practices, from remediation to the mitigation of plastic pollution.

5. <u>Conclusion</u>

In summary, the introduction of the Fluorescence Imaging Microplastic Analysis Platform (FIMAP) represents a significant advancement in the detection and identification of MPs. FIMAP enables simultaneous detection and classification of NR-stained MPs using machine learning techniques, incorporating a comprehensive process that includes fluorescence imaging under various excitation wavelengths and optical filters. By employing K-means clustering and spectral fingerprint comparison, FIMAP effectively detects and classifies ten of the most common MPs (PP, HDPE, LDPE, PS, EPS, PC, PVC, PET, PA), distinguishing them from common natural organic matter such as cotton, chitin, and wood. Additionally, our evaluation of FIMAP's adaptability for smaller MPs (35–100μm) and its ability to detect suspected MPs (>24.6μm) in sludge samples has identified potential limitations, highlighting areas for further refinement. Despite these challenges, this work underscores the significant progress made in MP analysis, offering a cost-effective, high-throughput, and objective method for classifying prevalent MPs. FIMAP, priced at under $3,000, emerges as a promising tool with broad environmental applicability, advancing our understanding of MP pollution and providing valuable insights for environmental policymakers and researchers alike.

## Abbreviations

HDPE: High-density polyethylene; LDPE: low-density polyethylene, PP: polypropylene; PS: polystyrene, EPS: Expanded polystyrene; ABS: acrylonitrile butadiene styrene; PC: polycarbonate; PVC: polyvinylchloride; PET: polyethylene terephthalate; PA: Polyamide; NR: Nile Red; MPs: Microplastics; FI: Fluorescence intensity; NOM: Natural organic matter; FIMAP: Fluorescence Imaging Microplastic Analysis Platform; ROI: Region of Interest.

## Authors Agreement

All authors have seen and approved the final version of the manuscript being submitted

## Declaration of Interests

There are no conflicts of interest to declare.

## Funding source

This work was supported by the USDA NIFA Hatch Formula Funds (Project # WIS03059).

## Data Availability

The code and data used in this manuscript are provided in our GitHub page: https://github.com/Isaac0047/Microplastic_Detection. This page includes the entire code for microplastic segmentation and classification. A detailed description of how to use the code is provided in the readme file.

## Acknowledgements

We thank Dr. K.G. Karthikeyan for securing the funding for this project and Dr. Pavana Prabhakar for her support in this project.

Shedding light on the Polymer's Identity: Microplastic Detection and Identification through Nile Red Staining and Multispectral Imaging (FIMAP)

By Derek Ho (dkho@upenn.edu)[1*] & Haotian Feng (haotian.feng@ucsf.edu)[2]

1 – Department of Mechanical Engineering and Applied Mechanics, University of Pennsylvania, Philadelphia, Pennsylvania 19103, United States;
Department of Biological Systems Engineering, University of Wisconsin-Madison, Madison, Wisconsin 53706, United States;
2 – Dept. of Radiation Oncology, University of California-San Francisco, San Francisco;
Department of Mechanical Engineering, University of Wisconsin-Madison
* Corresponding author

Figure S1: NR-staining method after Fenton-oxidation

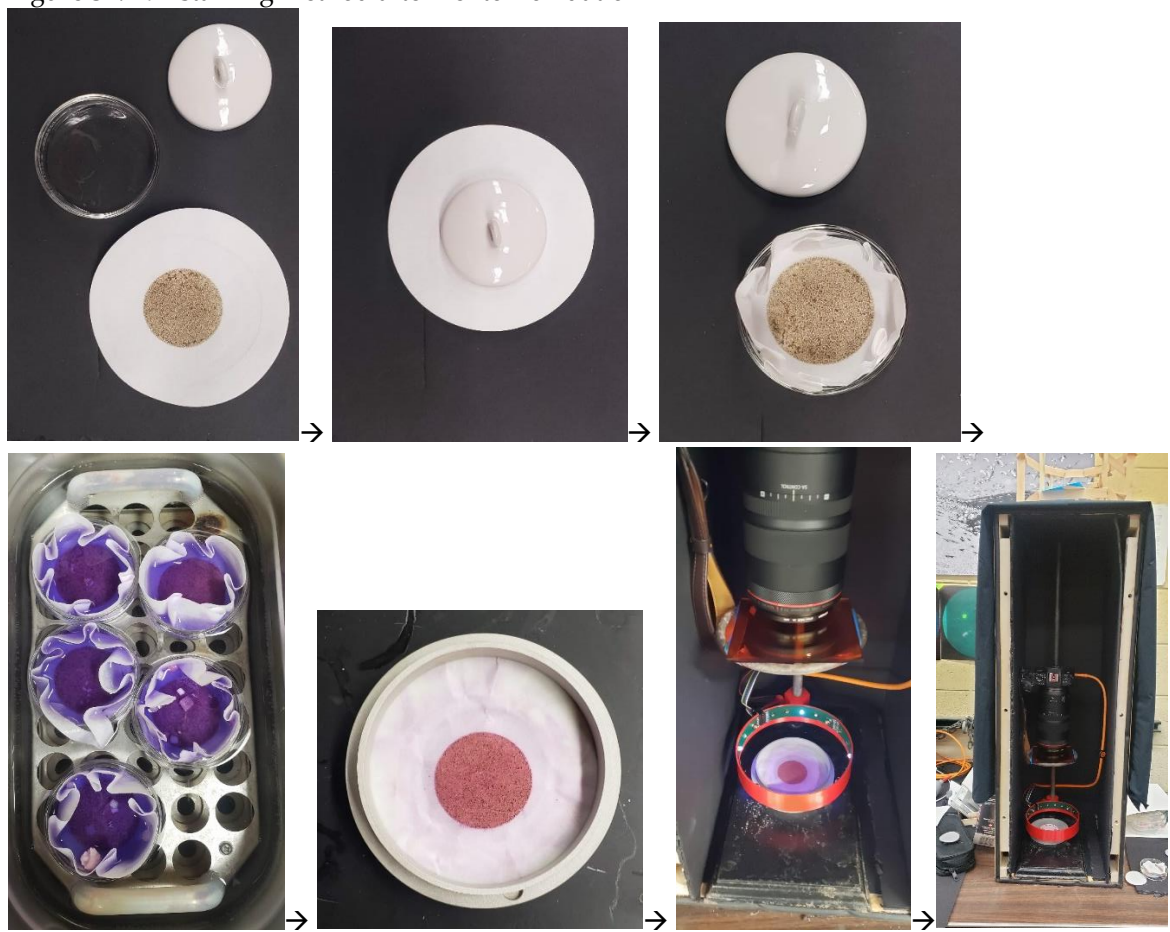

After filtering Fenton-oxidized sample onto an oversized 90mm filter, a crucible lid was used to stamp the filter paper into a 60mm diameter glass petri-dish. Here 15 ml of of NR solution [10 μg/mL] in acetone/water (25/75) was added and placed into a water bath (70°C) for 30 minutes with periodic agitation every 10 minutes. Upon staining, the filter paper was removed and analyzed using FIMAP.

Table S2: FIMAP and camera calibration settings

| FIMAP Illumination | | Camera settings | | Luminous exposure value (Lx.s) * |
|---|---|---|---|---|
| Wavelength (nm) | Lux (lx) | Exp time (s) | Absolute EV ±1 | |
| 265 (QT Brightek, QBHP684E-UV265N), | 1.0 | 4 | **2** | **4** |
| 305 (SETi/Seoul Viosys CUD1GF1A) | 0.3 | 15 | **~0** | **4.5** |
| 365 nm (Everlight Electronics Co Ltd, P6070U23240500-VD1M) | 1.5 | 2 | **3** | **3** |
| 405 (Everlight Electronics Co Ltd, P0010U23240500-VD1M) | 3.0 | 2 | **3** | **6** |
| 450 (Marktech Optoelectronics, MTE4600L-HP) | 70.3 | 2 | **3** | **140.6** |
| | Measured with luminosity meter | 1, 2, 4, 8, 15 [I stop apart] | | *Absorption spectrum may require more blue light to be absorbed than UV |

Absolute Exposure Value = $\log_2\left(\frac{N^2}{t}\right) + \log_2\left(\frac{ISO}{100}\right)$ [Unitless]     (Equation 2-1)

Luminous Exposure Value = lux × t     [lux.s]     (Equation 2-2)

Where  N = Aperture
          t = shutter speed
          lux = luminosity

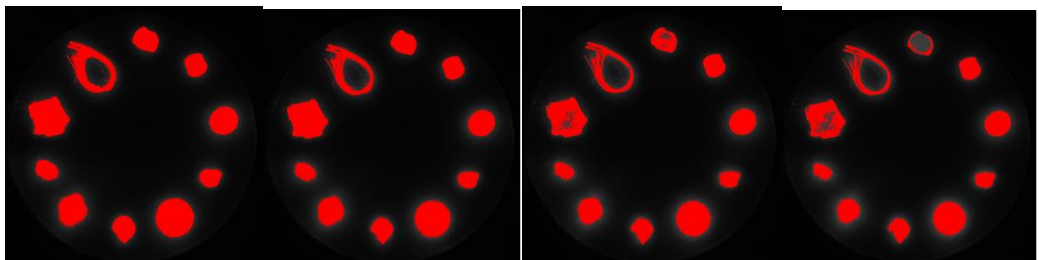

**Figure S2:** Image segmentation using ImageJ at thresholds of 10% [48,255], 9% [54,255], 8% [61,255], 7% [65,255], respectively

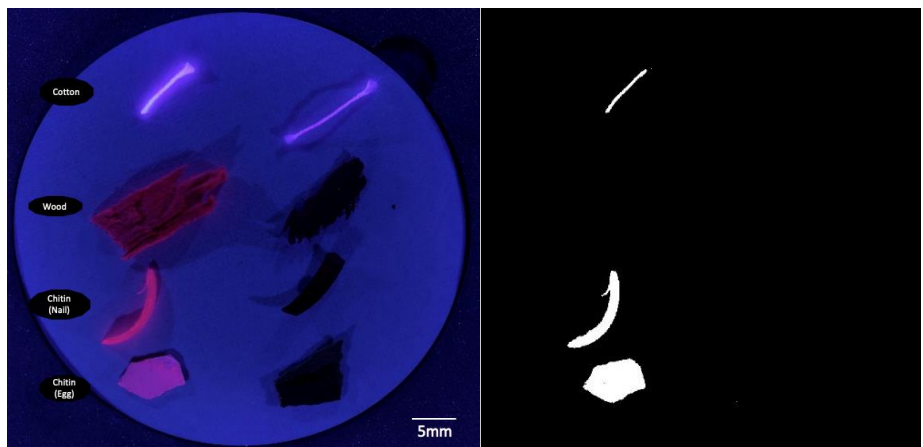

Figure S3: (left) NR-stained NOM of cotton (from top), wood, chitin-Nail and chitin-Egg. Control NR-NOM on the left and Fenton-treated NR-stained NOM on the right, captured at 265nm with no filter; (right) K-means clustering applied to NR-stained NOM in Fig 2 (left), exhibiting only detection of non Fenton-pretreated cotton, chitin-nail and chitin-egg, with all Fenton-treated NOM not detected.

Table S2: Segmentation results of MPs across 20 different filters

Box plot of areas of individual MPs captured across 20 different lighting/filter conditions of Table S2

| Mask | Highest # of ROI detected | Excludes | Incomplete ROI | Area of MPs (Pixels) | | | | | | | | | |
|---|---|---|---|---|---|---|---|---|---|---|---|---|---|
| | | | | PP | HDPE | LDPE | EPS | PS | ABS | PC | PVC | PET | PA |
| 1 | 7 | PP, HDPE, LDPE | | 114920 | 88242 | 135220 | 5151 | 155740 | 94417 | 50501 | 85882 | 135450 | 262260 |
| 2 | 7 | PP, HDPE, LDPE | | 115660 | 92966 | 132900 | 11843 | 176860 | 114610 | 50270 | 74297 | 104350 | 190230 |
| 3 | 7 | PP, HDPE, LDPE | PC, PET | 114410 | 89241 | 135280 | 18289 | 188630 | 131630 | 36919 | 63423 | 2848 | 136360 |
| 4 | 3 | EPS, PS, ABS, PC, PVC, PET, PA | | 116270 | 90006 | 134780 | 66786 | 313170 | 227490 | 79708 | 170260 | 1797 | 253190 |
| 5 | 3 | EPS, PS, ABS, PC, PVC, PET, PA | | 116660 | 92220 | 132360 | **49222** | 158140 | 79656 | 49100 | 61092 | 11160 | 20980 |
| 6 | 3 | EPS, PS, ABS, PC, PVC, PET, PA | | 138080 | 92883 | 147550 | 36451 | 249120 | 175130 | 63222 | 168110 | 163020 | 481040 |
| 7 | 4 | EPS, ABS, PC, PVC, PET, PA | | 166640 | 114560 | 195330 | 41119 | 148110 | 273160 | 68525 | 178220 | 172380 | 421080 |
| 8 | 4 | EPS, ABS, PC, PVC, PET, PA | | 139330 | 94326 | 189420 | 46134 | 158710 | 338780 | 61837 | 146550 | 152590 | 337050 |
| 9 | 4 | EPS, ABS, PC, PVC, PET, PA | | 149560 | 95245 | 197280 | 48251 | 143470 | 226570 | 70782 | 162500 | 147390 | 198140 |
| 10 | 7 | PP, HDPE, LDPE | | 174780 | 110760 | 219520 | **52826** | 290840 | 188290 | 68391 | 146640 | 154770 | 181060 |
| 11 | 10 | - | HDPE | 112380 | 54511 | 138930 | 9203 | 166600 | 94643 | 51757 | 85422 | 132380 | 258710 |
| 12 | 10 | - | | 116110 | 93838 | 141770 | 16576 | 209070 | 111290 | 52005 | 72725 | 120270 | 183170 |
| 13 | 4 | HDPE, EPS, PC, PVC, PET, PA | | 108880 | 89870 | 134500 | 43323 | 186830 | 79958 | 58070 | 146960 | 156920 | 348180 |
| 14 | 8 | PET, PA | HDPE | 112670 | 59000 | 137570 | 20862 | 180690 | 90458 | 49579 | 62821 | 148010 | 251550 |
| 15 | 9 | PA | | 118180 | 93293 | 150850 | 34504 | 177730 | 87186 | 51266 | 62946 | 107680 | 292910 |
| 16 | 7 | PP, HDPE, EPS | | 122510 | 93625 | 109300 | 37534 | 148150 | 94578 | 49412 | 86584 | 126140 | 252130 |
| 17 | 10 | (HDPE) | HDPE | 107970 | 68460 | 133690 | 12882 | 207220 | 140390 | 52162 | 85750 | 129090 | 214690 |
| 18 | 8 | PP, HDPE | PET | 129130 | 94846 | 124950 | 13175 | 212550 | 152170 | 37294 | 65643 | 61087 | 145270 |
| 19 | 9 | HDPE | PET, PA | 100820 | 96141 | 131300 | 24631 | 187090 | 111280 | 51628 | 78880 | 60981 | 42171 |
| 20 | 10 | - | PA | 111850 | 87799 | 136980 | 36824 | 194490 | 108470 | 52989 | 84014 | 139640 | 53226 |
| | | Ground truth of area – Median values (*1st Quartile values) | | 116190 | 92925 | 136130 | 35477.5 | 183760 | *94458 | 51881 | *674134 | 130735 | 233120 |
| | | Intersection over Union (IOU) | | 1.00 | 0.99 | 0.96 | 0.47 | 0.88 | 0.85 | 1.00 | 0.93 | 0.92 | 0.79 |

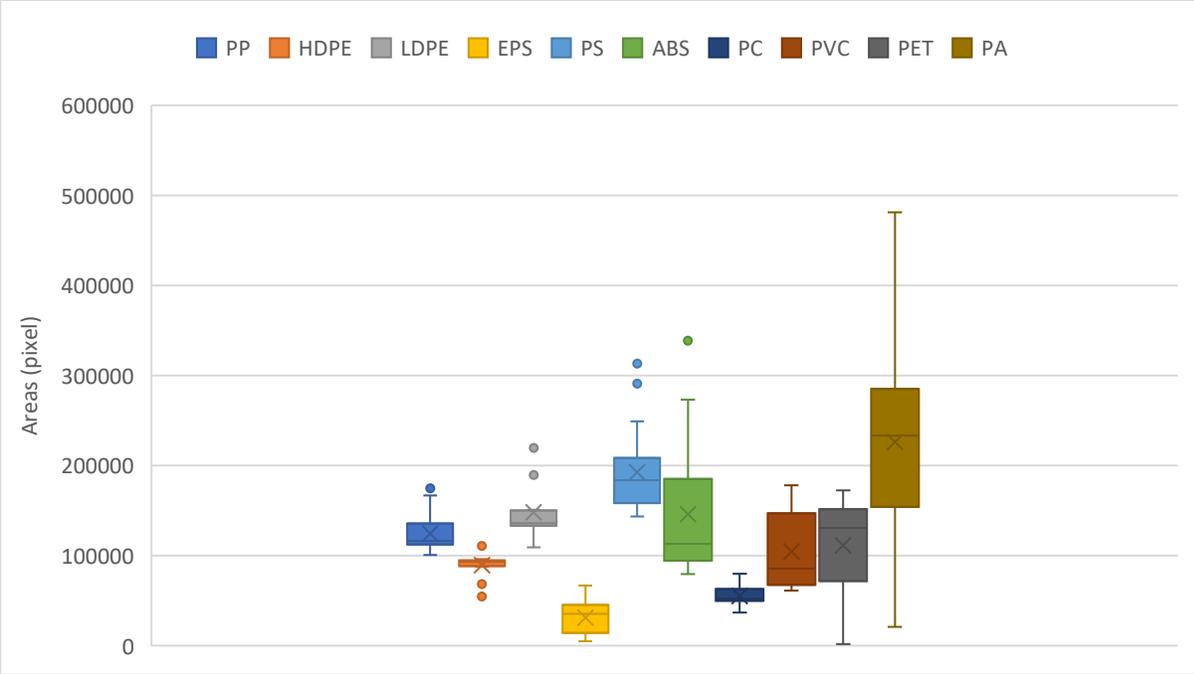